\documentclass{ieeeaccess}
\usepackage{cite}
\usepackage{amsmath}
\usepackage{algorithmic}
\usepackage{hyperref}
\usepackage[nameinlink]{cleveref}
\usepackage{graphicx}
\usepackage{caption}
\usepackage{multirow}
\usepackage[running]{lineno}
\usepackage{booktabs}
\usepackage{colortbl}
\usepackage{cuted}

\def\BibTeX{{\rm B\kern-.05em{\sc i\kern-.025em b}\kern-.08em
    T\kern-.1667em\lower.7ex\hbox{E}\kern-.125emX}}
\begin{document}
\history{Date of publication xxxx 00, 0000, date of current version xxxx 00, 0000.}
\doi{10.1109/ACCESS.2017.DOI}

\title{TransDAE: Dual Attention Mechanism in a Hierarchical Transformer for Efficient Medical Image Segmentation}
\author {\uppercase{Bobby Azad}\authorrefmark{1}
,
\uppercase{Pourya Adibfar}\authorrefmark{2}, Kaiqun Fu \authorrefmark{3},
}
\address[1]{{Department of Computer Science, South Dakota State University, South Dakota, United States. (e-mail: Bobby.Azad@jacks.sdstate.edu)}
\address[2]{Department of Computer Engineering, Technical and Vocational University (TVU), Shiraz, Iran (e-mail: pouriya.adibfar@gmail.com)}
\address[3]{Department of Computer Science, South Dakota State University, South Dakota, United States. (e-mail: kaiqun.fu@sdstate.edu)}}

\markboth
{Author \headeretal: Preparation of Papers for IEEE TRANSACTIONS and JOURNALS}
{Author \headeretal: Preparation of Papers for IEEE TRANSACTIONS and JOURNALS}

\corresp{Corresponding author: Bobby Azad (e-mail: Bobby.Azad@jacks.sdstate.edu).}

\begin{abstract}
In healthcare, medical image segmentation is crucial for accurate disease diagnosis and the development of effective treatment strategies. Early detection can significantly aid in managing diseases and potentially prevent their progression. Machine learning, particularly deep convolutional neural networks, has emerged as a promising approach to addressing segmentation challenges. Traditional methods like U-Net use encoding blocks for local representation modeling and decoding blocks to uncover semantic relationships. However, these models often struggle with multi-scale objects exhibiting significant variations in texture and shape, and they frequently fail to capture long-range dependencies in the input data. Transformers designed for sequence-to-sequence predictions have been proposed as alternatives, utilizing global self-attention mechanisms. Yet, they can sometimes lack precise localization due to insufficient granular details. To overcome these limitations, we introduce TransDAE: a novel approach that reimagines the self-attention mechanism to include both spatial and channel-wise associations across the entire feature space, while maintaining computational efficiency. Additionally, TransDAE enhances the skip connection pathway with an inter-scale interaction module, promoting feature reuse and improving localization accuracy. Remarkably, TransDAE outperforms existing state-of-the-art methods on the Synaps multi-organ dataset, even without relying on pre-trained weights.
\end{abstract}

\begin{keywords}
Transformer, Semantic segmentation, Deep learning, Medical image analysis
\end{keywords}

\titlepgskip=-15pt

\maketitle

\section{Introduction}
\label{sec:introduction}
\PARstart{E}{arly} disease diagnosis is paramount in healthcare, as it enables the detection of disorder severity and spread at initial stages \cite{gupta2023segpc}. Medical image segmentation serves as a critical component in automating computer-aided disease diagnosis (CAD), treatment planning, and surgical pre-assessment. The segmentation process involves partitioning target organ and tissue shapes and volumes through pixel-wise classification \cite{wu2022d}. Traditional manual annotation is labor-intensive, time-consuming, and susceptible to human error \cite{zhang2021pyramid}. Consequently, automating medical image segmentation has become a research focus to alleviate this burden. Recent studies have explored the potential of deep learning in CAD, given its successful application in various medical fields.

Over the years, convolutional neural networks (CNNs) and fully convolutional networks (FCNs) have been the dominant architectures in medical image segmentation, largely due to their ability to learn hierarchical features through convolutional operations. Among these, U-Net has emerged as a particularly effective model due to its U-shaped structure with symmetric encoding and decoding paths \cite{ronneberger2015u}. This architecture, with its skip connections, facilitates the fusion of low-level and high-level features, improving context modeling and producing accurate segmentation results. Variants such as Res-UNet \cite{diakogiannis2020resunet}, Dense-UNet \cite{cao2020denseunet}, U-Net++ \cite{zhou2018unet++}, and UNet3+ \cite{huang2020unet} have further improved performance by addressing specific limitations, such as the loss of spatial information in deeper layers.

However, despite these advances, CNN-based models face inherent limitations. The localized nature of convolution operations means that these models often struggle with capturing long-range dependencies and multi-scale variations within medical images, which are crucial for segmenting complex anatomical structures. Although efforts such as atrous convolution \cite{yu2015multi} and pyramid pooling \cite{zhao2017pyramid} have aimed to capture larger contextual information, the challenge of modeling global relationships within medical images remains. As a result, attention mechanisms have been integrated into CNN architectures to enhance their ability to focus on important regions, as demonstrated by models like Attention U-Net \cite{oktay2018attention} and its variants \cite{alryalat2022deep}.

In recent years, the introduction of Transformer architectures has provided a new avenue for addressing these challenges. Originally developed for natural language processing (NLP) \cite{vaswani2017attention}, Transformers are designed to capture long-range dependencies through self-attention mechanisms. This capability has inspired their application to computer vision tasks, including medical image segmentation. The Vision Transformer (ViT) \cite{dosovitskiy2020image}, for example, has demonstrated that self-attention can be effectively applied to image patches, achieving impressive results in various image recognition tasks. However, ViT and similar Transformer models come with their own set of challenges. The quadratic computational complexity of self-attention, coupled with the large amount of training data required, can make these models impractical for high-resolution medical images. Moreover, while Transformers excel at modeling global dependencies, they often lack the ability to capture fine-grained, local details, which are essential for accurate segmentation \cite{azad2023laplacian}.

To address these limitations, we propose TransDAE, a novel hierarchical Transformer model specifically designed for medical image segmentation. Our approach reimagines the self-attention mechanism by integrating both spatial and channel-wise associations across the entire feature space. This dual attention mechanism allows our model to maintain computational efficiency while capturing both local and global dependencies more effectively. Furthermore, we introduce an Inter-Scale Interaction Module (ISIM) to enhance the skip connection pathway, promoting feature reusability and improving localization accuracy. This module plays a crucial role in ensuring that the model can handle the multi-scale nature of medical images, which is often a challenge for both CNNs and Transformers.
In this work, our contributions can be summarized as follows:

\begin{itemize}
\item We propose a dual attention mechanism that simultaneously captures spatial and channel-wise dependencies, addressing the limitations of existing methods that primarily focus on one or the other.

\item We incorporate efficient self-attention and enhanced self-attention mechanisms to reduce computational complexity while effectively modeling both local and global dependencies, making our model scalable to high-resolution medical images.

\item We emphasize the importance of skip connections in bridging the encoder and decoder components, integrating an Inter-Scale Interaction Module (ISIM) to enhance feature reuse and improve localization accuracy.

\item We integrate a large-kernel attention module that further enhances the information conveyed through skip connections, amplifying the effectiveness of low-level localization information and resulting in a more robust and efficient network.
\end{itemize}

\section{Literature review}
\label{sec:literature}

\subsection{CNN-based Segmentation Networks}
In recent years, deep learning methods have gained prominence in image segmentation tasks, replacing hand-crafted-feature based machine learning approaches. CNNs have become a popular choice for various medical image segmentation tasks, primarily due to the success of U-Net \cite{azad2022medical}. The U-shaped symmetric structure of U-Net incorporates skip connections between each block of the encoder and decoder, enabling the concatenation of higher-resolution feature maps from the encoder network with upsampled features for more accurate representations. The success of U-Net has inspired researchers to adapt its architecture and improve its performance by applying various strategies, such as Res-UNet \cite{diakogiannis2020resunet}, Dense-UNet \cite{cao2020denseunet}, U-Net++ \cite{zhou2018unet++}, and UNet3+ \cite{huang2020unet}. The 3D U-net \cite{cciccek20163d} was proposed as an enhanced version of U-Net, specifically designed for volumetric segmentation in three dimensions.

Oktay et al. \cite{oktay2018attention} introduced attention gates to U-Net's skip connections, emphasizing the importance of specific objects by focusing on critical objects and disregarding irrelevant ones. Alryalat et al. \cite{alryalat2022deep} employed a dual-attention strategy in U-Net skip connections to make the network concentrate on more representative channels using channel attention and to identify the most informative spatial regions in images using spatial attention. Zhou et al. developed U-Net++ \cite{zhou2018unet++} and demonstrated that using nested and dense skip connections to inject encoder feature maps into the decoder, rather than directly fetching them, improves network performance. However, due to the limited receptive field size of convolution operations, CNN-based methods primarily capture local dependencies and struggle to represent long-range dependencies. Although the dimensional size of images changes in different network blocks to cover a varying range, the operation remains limited to local information and not global contexts. The locality of convolution operations and their weight-sharing property hinder them from capturing global contexts.

To overcome the limitations of CNN networks, various approaches have been developed in recent years. Yu et al. \cite{yu2015multi} attempted to expand the receptive field size to capture global contexts without downsampling images and losing resolution, by employing atrous convolution with a dilation rate. Zhao et al. \cite{zhao2017pyramid} used pyramid pooling at different feature scales to model global information. Wang et al. \cite{wang2018non} proposed a non-local network to capture long-range dependencies by calculating response at a position as a weighted sum of all features within the input feature maps. Some studies, \cite{fu2019dual,alryalat2022deep}, have discovered self-attention modules' potential to address the deficiency of CNNs in long-range dependency modeling. Despite these efforts to mitigate the shortcomings of CNNs, they remain unable to fully satisfy clinical application requirements, as strong long-range dependencies exist in the data of these applications.

\subsection{Transformers}
The success of Transformer methods in natural language processing, where high dependency exists between words in text, has encouraged researchers to leverage these models' long-range dependency capabilities for image segmentation and recognition tasks. ViT \cite{dosovitskiy2020image} served as a foundational method, introducing Transformer approaches to machine vision and outperforming traditional CNN-based architectures. This method partitions input images into segments called patches and embeds each patch's location so that the network can consider the spatial dependence between patches. These patches are then fed into a Transformer encoder, which employs multi-head attention modules, followed by a multi-layer perceptron for classification.

To improve the performance of this novel approach, several enhanced versions of ViT have been proposed, including Swin Transformer \cite{liu2021swin}, LeViT \cite{graham2021levit}, and Twins \cite{chu2021twins}. Given the complexity of these models, the Swin Transformer \cite{liu2021swin} sought to reduce the number of model parameters by dividing image patches into windows and applying the Transformer exclusively within patches inside each window. An additional step was suggested to allow adjacent windows to interact with each other, based on the fundamental principle of CNNs: shifting the window and then reapplying the Transformer module.

Although Transformer-based methods have demonstrated great success in various domains, they are not without their limitations. One notable shortcoming is their weakness in capturing local information representation. Unlike CNNs, which inherently focus on local features due to their convolution operations, Transformers primarily excel at capturing long-range dependencies. This limitation can lead to suboptimal performance in tasks where local information plays a crucial role in understanding and processing the data. Consequently, it has become imperative to explore hybrid models that can effectively leverage the strengths of both CNNs and Transformers in order to address these limitations and improve overall performance.

\subsection{Hybrid CNN-Transformer Approaches}
Recent advances in medical image segmentation have sought to harness the strengths of both CNNs and Transformer architectures by incorporating Transformer layers into the encoder component of CNN networks. This enables the combined model to capture local information while also effectively modeling long-range dependencies. TransUNet \cite{chen2021transunet} serves as a pioneering approach in this regard, utilizing a ResNet-50 backbone to generate low-resolution feature maps, which are then encoded using a ViT model. The encoded features are subsequently upsampled via cascaded upsampling layers to produce the final segmentation map. However, integrating a pure Transformer-based model alongside a CNN model can increase network complexity by up to eight times. To address this challenge, Cao et al. \cite{cao2021swin} introduced Swin-UNet, which computes attention within a fixed window (analogous to the Swin-Transformer approach). As an added feature, Swin-UNet includes a patch-expanding layer that reshapes adjacent feature maps into higher-resolution feature maps during the upsampling process. In another related approach, Wu et al. \cite{wu2022fat} incorporated a Transformer module into the encoding layers by replacing the single-branch encoder with a dual encoder containing both CNN and Transformer branches. Furthermore, the researchers devised a feature adaptation module (FAM) and a memory-efficient decoder to overcome the computational inefficiency associated with fusing these branches and the decoding component. In a similar vein, Azad et al. \cite{azad2023improving} tackled the limitations of traditional CNN-based methods by introducing a "Context Bridge". This feature merges the U-Net's local representation capability into a transformer model, overcoming issues in modeling long-range dependencies and handling diverse objects. Furthermore, they substituted the standard attention mechanism with an "Efficient Self-attention" strategy, simplifying the architecture without compromising performance.

CNN-based methods are proficient in capturing local information but struggle with modeling long-range dependencies essential for medical image analysis. In contrast, Transformer-based methods excel in long-range dependency representation but lack local information-capturing mechanisms. Therefore, our research aims to develop a method that combines the strengths of both models while maintaining acceptable network complexity. We propose a dual attention module for handling spatial input features and channel context, utilizing Wang et al.'s efficient self-attention method \cite{wang2021transbts} and enhanced self-attention module \cite{huang2021missformer} to minimize complexity. Our redesigned Transformer block is incorporated into a U-Net-like architecture, highlighting the significance of skip connections for improved performance and accurate feature reconstruction. By integrating a large kernel approach, we enhance information transfer, increase low-level localization information effectiveness, and ultimately strengthen the model's overall performance via better encoder-decoder communication.

\section{Proposed Method}
\begin{figure*}[h]
\centering
  \includegraphics[width=500px]{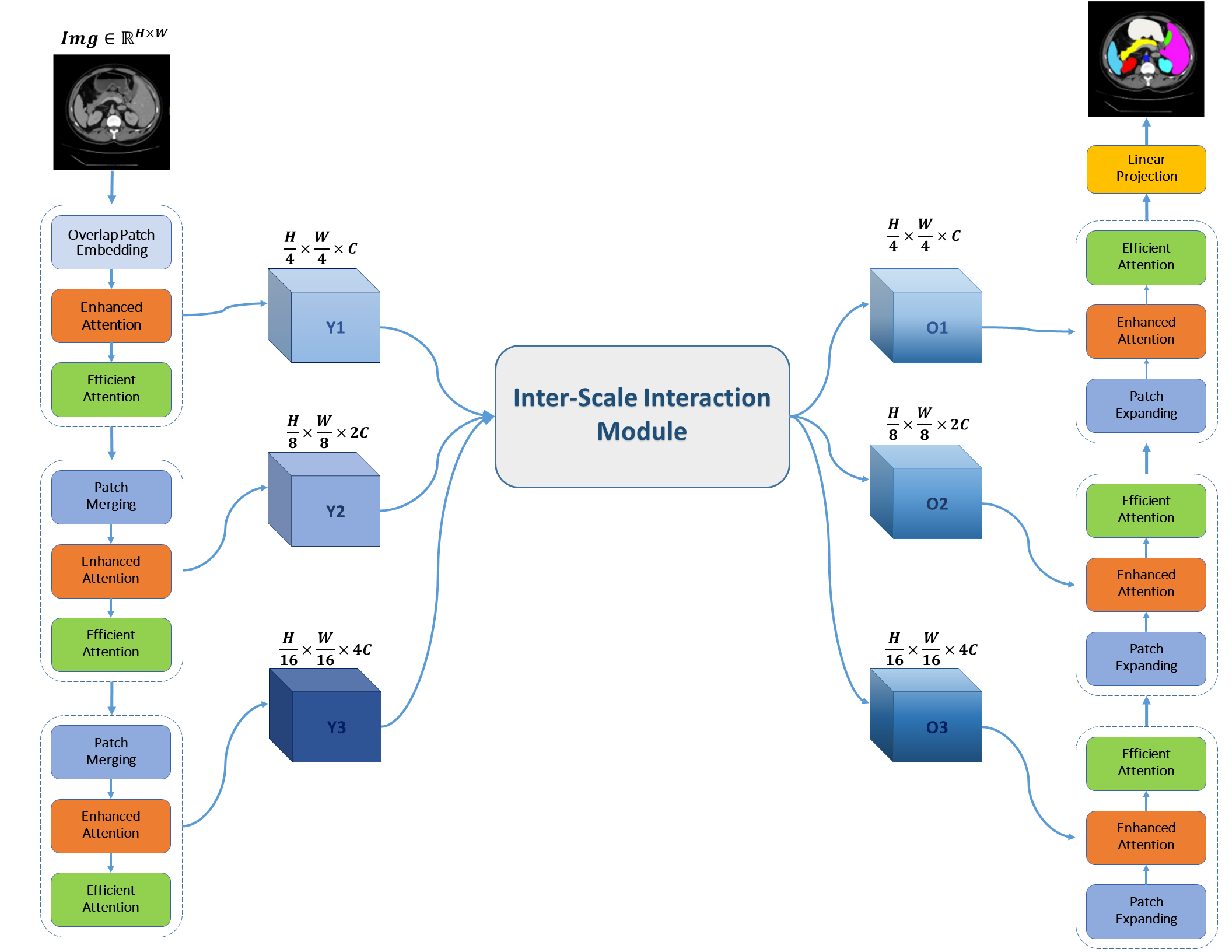}

  \caption{Overview of the Proposed Hierarchical Transformer Model. The model combines a U-Net-like structure with efficient dual attention mechanisms to achieve robust medical image segmentation. Starting with an input image \textbf{$x^{H\times W \times C}$}, the architecture tokenizes the input into overlapping patches. These tokens traverse through encoder modules that are made up of dual Transformer layers and patch merging functionality, enabling multi-scale hierarchical feature representation. During decoding, patch tokens are expanded and integrated with corresponding encoder features using a large-kernel attention module. This fusion process ensures better communication between the encoder and decoder components, with the final projection layer producing the output segmentation map.}
\label{fig:figure1}
\end{figure*}

\Cref{fig:figure1} presents an overview of our proposed model, a hierarchical Transformer model with a U-Net-like structure that leverages both local and global feature representation along with an enhanced skip connection module. Given an input image \textbf{$x^{H\times W \times C}$} with spatial dimensions $H \times W$ and $C$ channels, the model employs the patch embedding module\cite{cao2021swin, huang2021missformer} to obtain overlapping patch tokens of size $4 \times 4$. The tokenized input ($x^{n \times d}$) then passes through the encoder module, which comprises three stacked encoder blocks, each containing two sequential dual Transformer layers and a patch merging layer. Patch merging combines $2 \times 2$ patch tokens, reducing the spatial dimension while doubling the channel dimension, enabling the network to achieve a multi-scale representation hierarchically. In the decoder, tokens are expanded by a factor of $2$ in each block. The output from each patch-expanding layer is fused with features from the corresponding encoder layer's skip connection using an inter-scale interaction module. The fused features are processed by two sequential dual Transformer layers. Ultimately, a linear projection layer generates the output segmentation map. In the subsequent subsections, we will briefly discuss our dual attention and transformer block, followed by an introduction to our large-kernel attention module.

\subsection{Dual Attention Transformer Block}
\begin{figure*}[h]
\centering
  \includegraphics[width=500px]{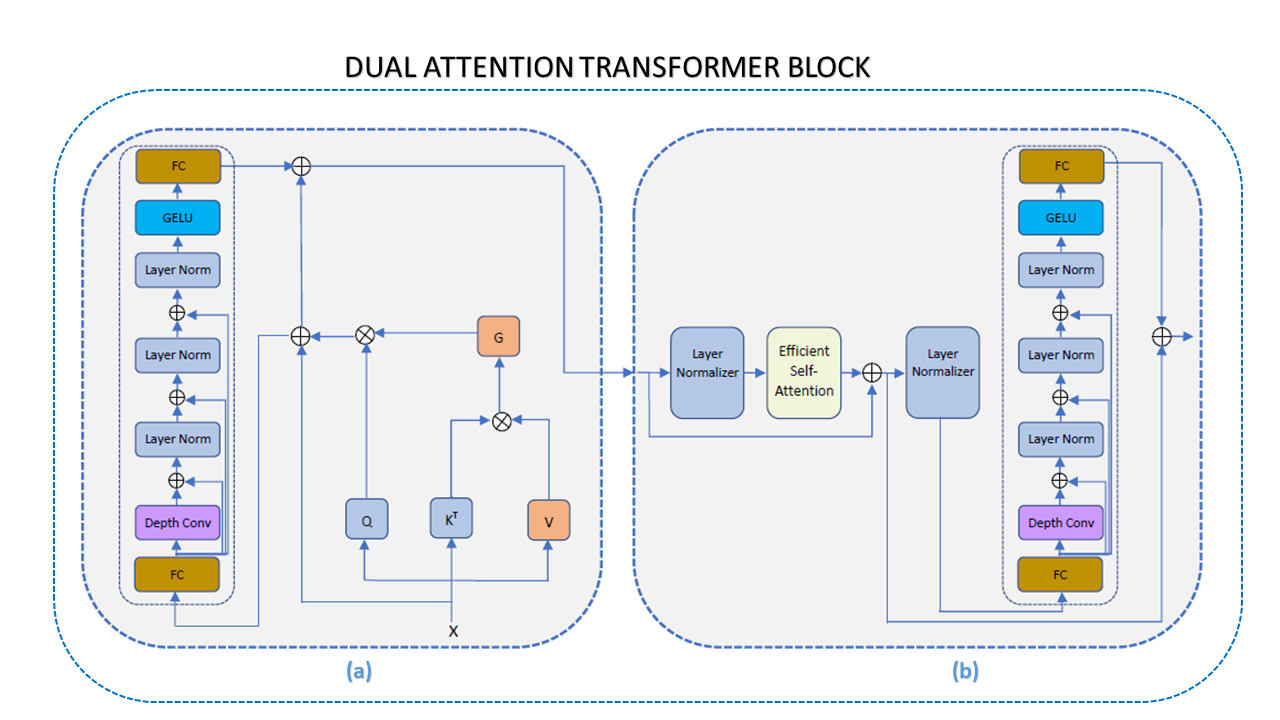}

  \caption{Visual depiction of the integrated dual attention mechanism. (a) Illustrates the channel attention process, emphasizing efficient channel-specific representations. (b) Portrays the spatial attention component, underscoring its ability to discern contextual dependencies within the image. Combined, these components work harmoniously to refine medical image segmentation by concentrating on both spatial relations and informative channels.}
\label{fig:figure2}
\end{figure*}

The motivation for incorporating dual attention in our model comes from the realization that both channel and spatial attention are essential in medical image segmentation tasks. Accurate segmentation results rely on the efficient representation of feature tensors. Channel attention enables the model to focus selectively on the most informative representation, fostering a deeper understanding of the structures within medical images. In contrast, spatial attention emphasizes spatial relationships between features, allowing the model to capture vital contextual information and dependencies across various regions in the image. By integrating dual attention, our model effectively combines the strengths of both channel and spatial attention, ultimately enhancing its performance in medical image segmentation tasks. This approach enables the development of a more robust and efficient network capable of representing feature tensors effectively, leading to improved segmentation outcomes. A visual representation of the dual attention mechanism is illustrated in \Cref{fig:figure2}. This figure illustrates how the channel and spatial attention components work in tandem to boost the model's segmentation capabilities. It is important to note that our design applies the attention mechanisms sequentially, not in parallel, leading to improved performance.

To harness the advantages of the dual attention mechanism without incurring its associated complexity, we use an efficient attention module for channel attention and an enhanced transformer block for spatial attention.
The limitation of the standard self-attention mechanism is its quadratic computational complexity ($O(N^2)$), as shown in Equation \eqref{eq:self_attention}. This restricts the architecture's applicability to high-resolution medical images.

\begin{equation}
S(\mathbf{Q}, \mathbf{K}, \mathbf{V}) =\operatorname{softmax}\left(\frac{\mathbf{Q K^T}}{\sqrt{d_{\mathbf{k}}}}\right) \mathbf{V}.\label{eq:self_attention}
\end{equation}

In Equation \eqref{eq:self_attention}, $Q$, $K$, and $V$ denote the query, key, and value vectors, respectively, while $d$ represents the embedding dimension. By adopting the efficient attention mechanism, we reduce the computational overhead without sacrificing the benefits offered by the channel attention approach. This allows our model to process feature maps more effectively and deliver enhanced performance in medical image segmentation tasks. Furthermore, the efficient attention mechanism ensures that the model remains scalable, enabling its application to a broader range of use cases and datasets. We utilize the Efficient Attention method proposed by Shen et al.~\cite{shen2021efficient} as Equation \eqref{eq:efficient-attention}:

\begin{align}
\label{eq:efficient-attention}
\mathbf{E}(\mathbf{Q},\mathbf{K},\mathbf{V}) = \mathbf{\rho_{q}}(\mathbf{Q})(\mathbf{\rho_{k}}(\mathbf{K})^{\mathbf{T}}\mathbf{V}),
\end{align}

Here, $\rho_{q}$ and $\rho_{k}$ represent normalization functions for queries and keys, respectively. As demonstrated by Shen et al.~\cite{shen2021efficient}, applying softmax normalization functions as $\rho_{q}$ and $\rho_{k}$ makes the module output equivalent to dot-product attention. Consequently, Efficient Attention first normalizes keys and queries, multiplies keys and values, and then multiplies the resulting global context vectors by the queries to produce a new representation.

Efficient Attention differs from dot-product attention in that it does not initially compute pairwise similarities between points. Instead, the keys are represented as $d$ attention maps $\mathbf{k^{T}}{j}$, where $j$ denotes the position $j$ in the input feature. These global attention maps reflect the semantic aspects of the entire input feature, rather than similarities to the input's position. This reordering significantly reduces the computational complexity of the attention mechanism while retaining a high representational capacity. With memory complexity at $O(dn + d^{2})$ and computational complexity at $O(d^{2}n)$ for typical settings ($d{v} = d, d_{k} = \frac{d}{2}$), our structure employs Efficient Attention to capture the channel-wise significance of the input feature map.

To reduce the complexity of the spatial attention module, we follow Huang et al.'s \cite{huang2021missformer} strategy, which is a spatial reduction self-attention that can be applied to high-resolution feature maps. In this strategy proposed by Huang et al., using the spatial reduction ratio R we allow the spatial resolution to be reduced so that self-attention can be achieved effectively. Equation \eqref{eq:enhance-attention} illustrates the mathematical formulation of this reduction strategy.

\begin{equation}
\label{eq:enhance-attention}
n e w_{-} K=\operatorname{Reshape}\left(\frac{N}{R}, C \cdot R\right) W(C \cdot R, C),
\end{equation}

As shown in the equation, first, $K$ and $V$ are reshaped to a new shape $\frac{N}{R} \times(C \cdot R)$. Then, using a linear projection $W$, channel depth restores to $C$. These operations reduce the complexity of self-attention to $\mathcal{O}\left(\frac{N^2}{R}\right)$, which is computationally viable for applying to high-resolution feature maps. For implementing spatial reduction, techniques such as convolution or pooling can be employed.

\subsection{Inter-scale Interaction Module}

The attention mechanism fundamentally serves as a dynamic selector, adept at emphasizing relevant features across scales while marginalizing redundant ones by relying on input features. An essential byproduct of this mechanism is the attention map, which operates akin to a spotlight, highlighting the relative significance of various features across different scales. This spotlight subsequently aids in deciphering how different features correlate.

In analyzing the diverse methodologies to establish relationships among features, two primary strategies emerge, each addressing different scales.

The first strategy utilizes what is commonly referred to as a "self-attention mechanism" \cite{dosovitskiy2020image,zhang2019self,azad2023enhancing}. While this mechanism excels in understanding long-distance dependencies, it exhibits certain limitations in the context of multiple scales:

\begin{itemize}
\item It naively processes images as 1D sequences, neglecting their inherent 2D structure.
\item Its computational demands are substantial, with its quadratic complexity being especially cumbersome for high-resolution images.
\item Despite its proficiency in spatial adaptability, it fails to adequately adapt across different scales and channels.
\end{itemize}

In contrast, the second strategy leverages the capabilities of large-kernel convolutions \cite{woo2018cbam,wang2017residual,park2018bam}. These convolutions are inherently adept at working across scales, discerning feature importance, and generating attention maps. This approach, however, is not without challenges. The primary concern is that introducing these large-kernel convolutions escalates both computational overheads and parameter counts.

\begin{figure}[h]
\begin{center}
  \includegraphics[width=\linewidth]{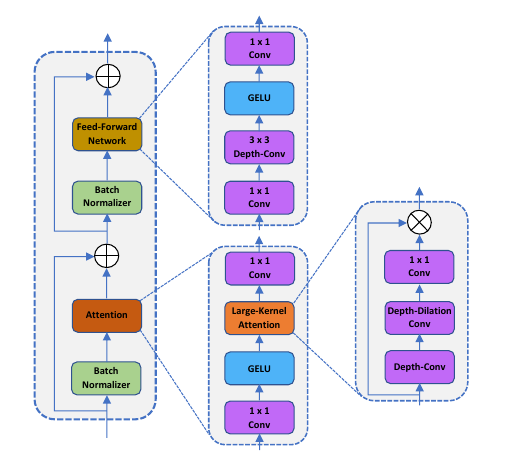}
\end{center}
  \caption{Schematic representation of the Inter-scale Interaction Module. This module skillfully integrates the benefits of both convolution and self-attention, circumventing the limitations of each. The module incorporates local context information, expansive receptive fields, linear complexity, and dynamic processes, ensuring adaptability across both spatial and channel dimensions. A central element of the module is the attention map, emphasizing the significance of each feature. The figure delineates the decomposition of large kernel convolution operations, capturing long-distance associations with reduced computational overhead and fewer parameters, a pivotal innovation of the Inter-scale Interaction Module.}
\label{fig:figure3}
\end{figure}

Drawing inspiration from the Visual Attention Network by Guo et al. \cite{guo2023visual}, we propose an innovative blend of both strategies: the self-attention mechanism and large-kernel convolutions. This blend addresses the challenge of interactions across different scales. Through the decomposition of large kernel convolution operations, we aim to gain a more intricate understanding of long-distance dependencies across scales. As depicted in \Cref{fig:figure3}, a large-kernel convolution can be astutely deconstructed into three principal segments addressing different scales:

\begin{itemize}
\item Spatial local convolution via depth-wise convolution.
\item Spatial long-range convolution through depth-wise dilation convolution.
\item Channel convolution, facilitated by a compact $1 \times 1$ convolution.
\end{itemize}

Upon further examination, a $K \times K$ convolution can be divided into three sub-elements addressing various scales: a $([k/2]*[k/2])$ depth-wise dilation convolution with dilation of $d$, an expanded $(2 d-1) \times(2 d-1)$ depth-wise convolution, and finally, a $1 \times 1$ convolution. This strategic segmentation is computationally efficient both in processing and parameterization. By identifying long-range relationships across scales, we are equipped to evaluate the prominence of individual points, culminating in our attention map's design.

Mathematically, our expansive inter-scale interaction module can be articulated as:

\begin{strip}
\begin{equation}
\begin{aligned}
\text{Attention} & = \operatorname{Conv}_{1 \times 1}(\mathrm{DW}-\mathrm{D}-\operatorname{Conv}(\mathrm{DW}-\mathrm{Conv}(\mathrm{F}))), \
\text{Output} & = \text{Attention} \otimes F .
\end{aligned}
\end{equation}
\end{strip}

Distinct from traditional attention methodologies, our inter-scale interaction strategy dispenses with auxiliary normalization functions such as sigmoid and softmax. We argue that the essence of attention methods does not reside in the normalization of attention maps, but in the adaptability of outputs based on input features across scales. By harmoniously integrating convolution and self-attention, our approach is holistic, considering local contexts, expansive receptive fields, linear complexity, and dynamism across scales. Given that different channels frequently correspond to unique objects within deep neural networks, this adaptability across channels becomes indispensable in visual tasks. Figure 3 represents the intricate details and structure of the Inter-scale Interaction Module.

\section{Experimental Results}
This section provides details about the training process, the metrics we used during our experimental evaluation, and a detailed analysis of the experimental results.

\subsection{Training process}
In this study, we implemented the proposed method using PyTorch on an NVIDIA Tesla V100 GPU with a 24-batch size without any data augmentation. For 400 epochs, we trained all the models at a learning rate of $1e-3$ and a decay rate of $1e-4$. The weight of the model was initialized using a standard normal distribution, which is stable from the start and ensures less fluctuation in weight. Furthermore, if the validation performance does not change in ten consecutive epochs during the training process, the training process stops. On both training and validation datasets, the optimization algorithm gradually decreased the loss value and eventually converged to the optimal solution during the training process. There was no evidence of instability during the training process.

\subsection{Dataset}
The proposed method was evaluated on Synapse multi-organ segmentation dataset \cite{landman2015miccai}. Beyond the Cranial Vault (BTCV) abdomen challenge, dataset \cite{landman2015miccai} includes 30 abdominal CT scans for a total of 3779 axial contrast-enhanced abdominal clinical CT images. Interpreters annotated 13 organs in each instance, including the spleen, the right kidney, the left kidney, the gallbladder, the esophagus, the liver, the stomach, the aorta, the inferior vena cava, the portal vein, the splenic vein, the pancreas, the left adrenal gland, and the right adrenal gland.  Each CT scan is acquired with contrast enhancement leading to volumes in the range of $85 \sim 198$ slices of $512 \times 512$ pixels.

\subsection{Quantitative and Qualitative Results}
Table 1 showcases a comparative analysis of our proposed approach against several benchmarking methods. This includes both our preliminary models (baselines) and a few top-performing, state-of-the-art architectures.

For a comprehensive understanding of our method's effectiveness, we evaluated three distinct baselines:

\noindent\textbf{Baseline:} This model forms the foundation of our approach and excludes the enhancements of dual attention and ISIM. Instead, it only employs an efficient attention module in each transformer block.

\noindent\textbf{Proposed Method (without ISIM):} An evolution of the initial model, this version incorporates both channel and spatial attention. As we describe it, this combines the efficiencies of both attention mechanisms.

\noindent\textbf{Proposed Method:} This embodies our holistic approach, utilizing all features, including ISIM.

Sequential enhancements in our model evidently enhanced its performance. Incorporating dual attention and subsequently, the ISIM, empirically validated our strategy's potency in addressing medical image segmentation challenges.

\definecolor{Proposed method}{RGB}{230,230,250} 

\begin{table*}[!ht]
    \centering
    \caption{A comparison of the proposed approach on the \textit{Synapse} dataset. \textcolor{blue}{Blue} highlights the leading result, while \textcolor{red}{red} represents the next best outcome.}
    \vspace{0.5em}
    \resizebox{\textwidth}{!}{
        \begin{tabular}{l|cc|cccccccc}
            \toprule
            \textbf{Methods}                        & \textbf{DSC~$\uparrow$} & \textbf{HD~$\downarrow$} & \textbf{Aorta}          & \textbf{Gallbladder}    & \textbf{Kidney(L)}      & \textbf{Kidney(R)}      & \textbf{Liver}          & \textbf{Pancreas}       & \textbf{Spleen}         & \textbf{Stomach}        \\
            \midrule
            \hline
            R50 U-Net \cite{chen2021transunet}      & 74.68                   & 36.87                    & 87.74                   & 63.66                   & 80.60                   & 78.19                   & 93.74                   & 56.90                   & 85.87                   & 74.16
            \\
            U-Net \cite{ronneberger2015unet}           & 76.85                   & 39.70                    & \textcolor{red}{89.07}  & \textcolor{red}{69.72}  & 77.77                   & 68.60                   & 93.43                   & 53.98                   & 86.67                   & 75.58
            \\
            R50 Att-UNet \cite{chen2021transunet}   & 75.57                   & 36.97                    & 55.92                   & 63.91                   & 79.20                   & 72.71                   & 93.56                   & 49.37                   & 87.19                   & 74.95
            \\
            Att-UNet \cite{schlemper2019attention}  & 77.77                   & 36.02                    & \textcolor{blue}{89.55} & 68.88                   & 77.98                   & 71.11                   & 93.57                   & 58.04                   & 87.30                   & 75.75
            \\
            R50 ViT \cite{chen2021transunet}        & 71.29                   & 32.87                    & 73.73                   & 55.13                   & 75.80                   & 72.20                   & 91.51                   & 45.99                   & 81.99                   & 73.95
            \\
            TransUNet \cite{chen2021transunet}      & 77.48                   & 31.69                    & 87.23                   & 63.13                   & 81.87                   & 77.02                   & 94.08                   & 55.86                   & 85.08                   & 75.62
            \\
            Swin-Unet \cite{cao2021swin}            & 79.13                   & 21.55                    & 85.47                   & 66.53                   & 83.28                   & 79.61                   & 94.29                   & 56.58                   & 90.66                   & 76.60
            \\
            LeVit-Unet \cite{xu2021levit}           & 78.53                   & \textcolor{red}{16.84}                    & 78.53                   & 62.23                   & 84.61                   & 80.25                   & 93.11                   & 59.07                   & 88.86                   & 72.76
            \\
            MT-UNet \cite{wang2022mixed} & 78.59 & 26.59 & 87.92 & 64.99 & 81.47 & 77.29 & 93.06 & 59.46 & 87.75 & 76.81 \\
            TransDeepLab \cite{azad2022transdeeplab} & 80.16 & 21.25 & 86.04 & 69.16 &84.08 & 79.88 & 93.53 & 61.19 & 89.00 & 78.40 \\
            FFUNet-trans \cite{xie2022ffunet}    & 79.09 & 31.65  & 86.68 & 67.09 & 81.13 & 73.73  & 93.67 & \textcolor{red}{64.17} & 90.92 & 75.32 \\
            HiFormer \cite{heidari2022hiformer}     & 80.39  & \textcolor{blue}{14.70}  & 86.21  & 65.69  & 85.23  & 79.77 & 94.61 & 59.52  & 90.99 & \textcolor{blue}{81.08}
            \\
            \hline
            \hline
            Baseline                              & 80.66                   & 17.00   & 85.81                   & 66.89                   & 84.40                   & 80.51 & \textcolor{red}{94.80}  & 62.25                   & 91.05                   & \textcolor{red}{79.58}
            \\
            
            Proposed Method (without ISIM) & \textcolor{red}{81.45} & 17.32  & 87.63  & 69.59  & \textcolor{red}{85.32} & \textcolor{red}{80.57} & 94.71 & 63.91  & \textcolor{red}{91.49} & 78.42
            \\
            \rowcolor{Proposed method}
                \textbf{Proposed method} & \textbf{\textcolor{blue}{82.16}} & \textbf{17.41} & \textbf{88.89} & \textbf{\textcolor{blue}{71.48}} & \textbf{\textcolor{blue}{85.45}} & \textbf{\textcolor{blue}{80.85}} & \textbf{\textcolor{blue}{94.85}} & \textbf{\textcolor{blue}{65.02}} & \textbf{\textcolor{blue}{91.62}} & \textbf{79.13}
            \\
            \bottomrule
        \end{tabular}
    }\label{tab:performance_comparison}
\end{table*}

\begin{figure}[h]
\begin{center}
  \includegraphics[width=\linewidth]{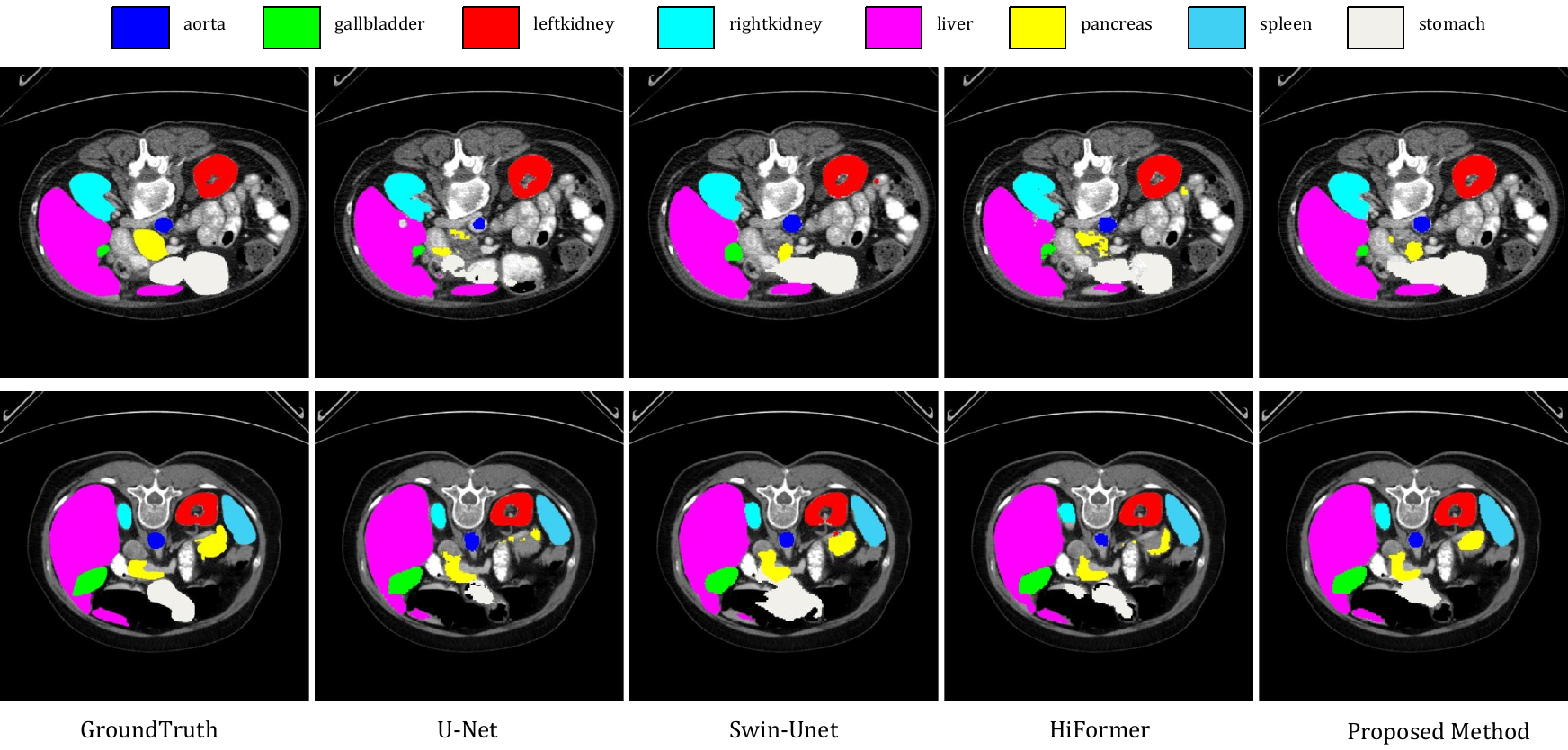}
\end{center}
  \caption{Segmentation comparisons on the \textit{Synapse} dataset reveal that our suggested approach produces more refined and smooth borders for the stomach, spleen, and liver organs while also displaying fewer false positive prediction masks for the gallbladder in comparison to Swin-Unet and HiFormer. In the bottom row, the proposed method additionally demonstrates a reduced false positive area for the pancreas.}
\label{fig:figure5}
\end{figure}

\begin{figure}[h]
\centering
\resizebox{0.48\textwidth}{!}{
    \begin{tabular}{@{} *{3}c @{}}
    \includegraphics[width=0.18\textwidth]{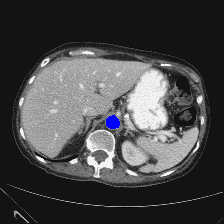} &
    \includegraphics[width=0.18\textwidth]{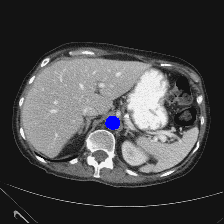} &
    \includegraphics[width=0.18\textwidth]{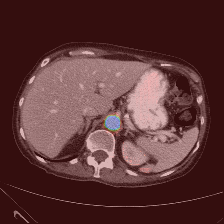} \\
    \includegraphics[width=0.18\textwidth]{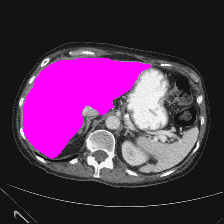} &
    \includegraphics[width=0.18\textwidth]{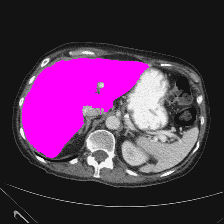} &
    \includegraphics[width=0.18\textwidth]{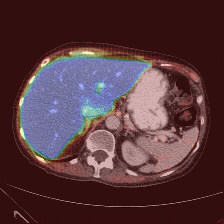} \\
    \includegraphics[width=0.18\textwidth]{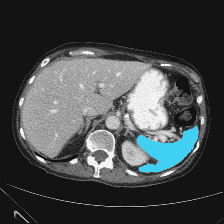} &
    \includegraphics[width=0.18\textwidth]{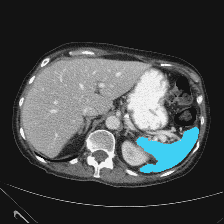} &
    \includegraphics[width=0.18\textwidth]{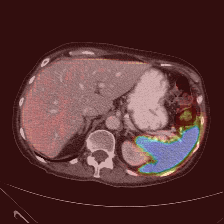}  \\
    \includegraphics[width=0.18\textwidth]{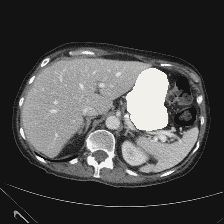} &
    \includegraphics[width=0.18\textwidth]{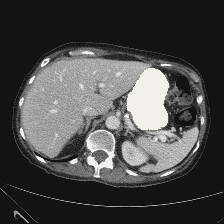} &
    \includegraphics[width=0.18\textwidth]{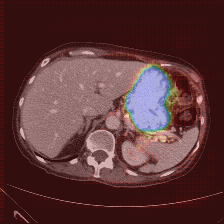}  \\
    {\small (a) Ground Truth} & {\small(b) Prediction} & {\small(c) Heatmap} 
    \end{tabular}
}
\caption{Visual representation of the attention map for the proposed model using Grad-CAM \cite{selvaraju2017grad} on the \textit{Synapse} dataset. The outcomes illustrate the efficiency of our approach in identifying large organs (liver, spleen, and stomach organs arranged from top to bottom), which demonstrates our method's proficiency in capturing long-range dependencies.} \label{fig:largeorgan}
\end{figure}

In this thorough comparison with top-tier models, our methodology underscores its superiority. To emphasize, the Dice Similarity Coefficient (DSC) of our proposal impressively settles at 82.16\%, outclassing formidable contenders like the HiFormer, which rests at 80.39\%.

A salient feature of our model is its aptitude for delineating finer anatomical structures. The integration of ISIM markedly amplifies this capability. This prowess is evident in the Gallbladder's segmentation, where our technique delivers a 71.48\% score, overtaking others like the TransDeepLab's 69.16\%. Similarly, the Pancreas, a traditionally intricate organ to segment due to its size, witnesses a conspicuous uplift with our method, achieving 65.02\%, surpassing even the FFUNet-trans's 64.17\%.

In the segmentation of more pronounced organs, our model remains unparalleled. The Kidney (L) and Kidney (R) respectively logged scores of 85.45\% and 80.85\%. Noteworthy is the Liver's segmentation, where our approach, with a score of 94.85\%, nearly mirrors the HiFormer's 94.61\%. Furthermore, in segmenting the Spleen, our model, at 91.62\%, slightly edges out our own baseline, which clocked 91.05\%.

\subsection{Ablation Study}
A defining trait of our technique is its ability to adeptly capture long-range dependencies. The superior prediction capabilities for larger organs, such as the liver, compared to other models, stand testament to this. The model's prowess in accommodating these long-range dependencies within its predictive realm is significant.

Additionally, we observed that for smaller organs, such as the aorta, U-Net models tend to outshine other Transformer-based methodologies. This highlights the indispensable role of local feature representation when predicting smaller entities and the consequential need to assimilate this information into the prediction matrix.

Reinforcing our point on the model's capability to harness long-range information, it's imperative to note our method's adeptness in segmenting both small and large organs. This demands a considerable receptive field size for precision in object prediction. We further elucidate this with a class activation map for both organ types in \Cref{fig:figure5}, shedding light on our model's enhanced ability to discern local patterns, resulting in meticulous segmentation.

\section{Conclusion}
In this study, we presented and assessed a new architecture designed for medical image segmentation, which harmoniously combines efficient and enhanced attention mechanisms and incorporates the unique capabilities of the ISIM. Our structured approach of evaluating the model through incremental baselines clearly highlighted the individual contributions of each component, with a special emphasis on the transformative role of the ISIM in boosting overall performance. Beyond outperforming our foundational models, our proposed method stood toe-to-toe with, and in many instances exceeded, the performance of top-tier contemporary architectures. Given its impressive accuracy and efficiency, our model holds significant clinical value, positioning itself as an invaluable aid for healthcare practitioners in diagnostic and therapeutic endeavors. This seamless fusion of groundbreaking research with tangible real-world implications not only accentuates the importance of our methodology but also sets a promising trajectory for future innovations in medical imaging.

\bibliographystyle{unsrt}
\bibliography{Ref}

\EOD

\end{document}